\documentclass[10pt, conference]{ieeeconf}

\IEEEoverridecommandlockouts
\title{\LARGE \bf Stance Control Inspired by Cerebellum Stabilizes Reflex-Based Locomotion on HyQ Robot}

\author{Gabriel Urbain$^{1}$, Victor Barasuol$^{2}$, Claudio Semini$^{2}$, Joni Dambre$^{1}$ and Francis wyffels$^{1}$
\thanks{$^{1}$ Gabriel Urbain, Joni Dambre and Francis wyffels are with IDLab-AIRO -- Ghent University -- imec, Ghent, Belgium. Corresponding email: {\tt\small gabriel.urbain@ugent.be}}%
\thanks{$^{2}$ Victor Barasuol and Claudio Semini are with Dynamic Legged Systems (DLS) Lab, Istituto Italiano di Tecnologia (IIT), Genova, Italy}
}

\usepackage{amsmath, amsmath, amssymb, lipsum, standalone, subfig, tikz}

\usetikzlibrary{arrows, backgrounds, fadings, decorations.markings, calc}
\begin{document}

\begin{figure*}[h]
  \vspace{-1.7cm}
  \hspace{-0.8cm}\includegraphics[width=1.1\textwidth]{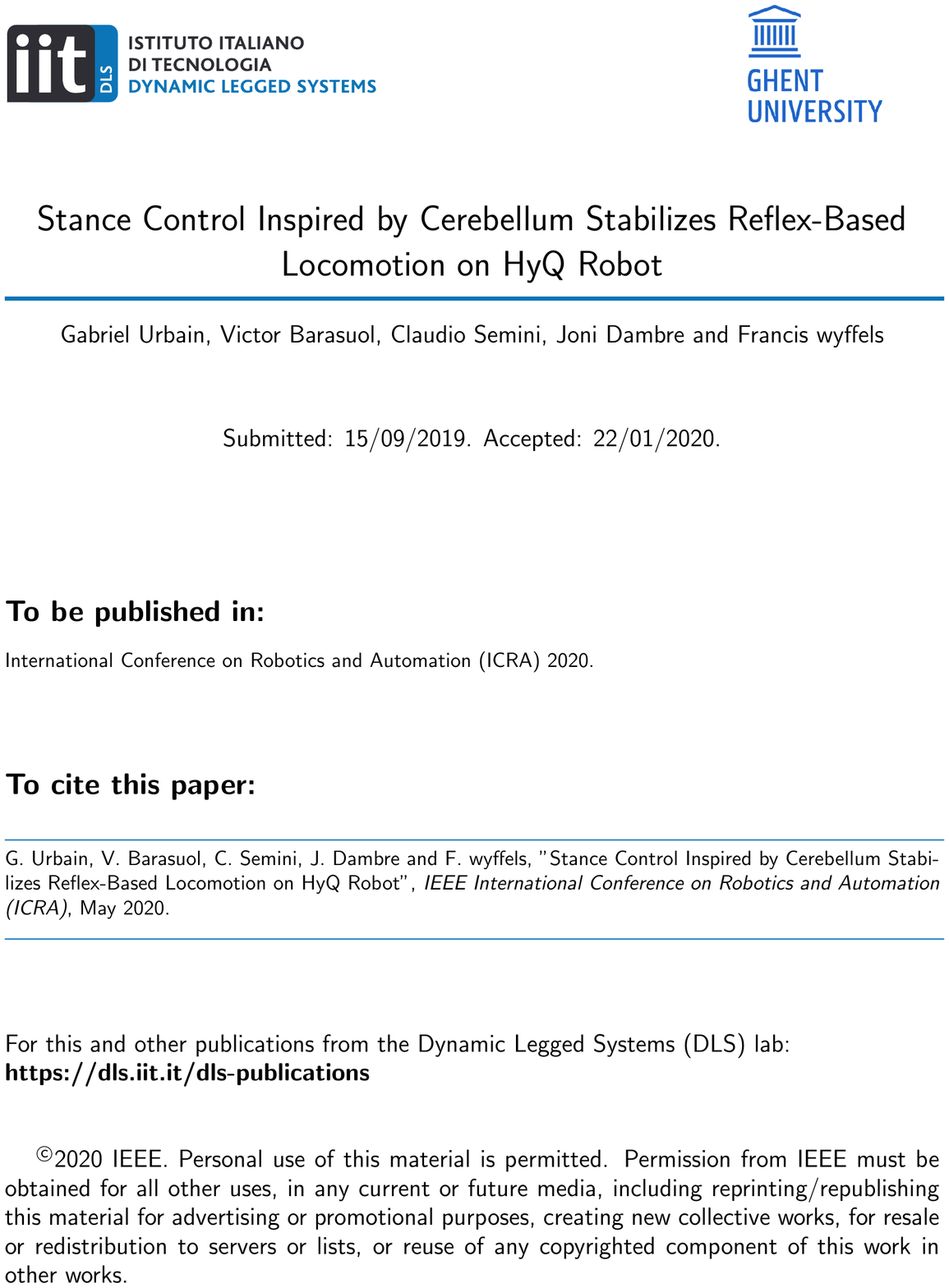}
\end{figure*}

\newpage



\maketitle
\thispagestyle{empty}
\pagestyle{empty}

\begin{abstract}
Advances in legged robotics are strongly rooted in animal observations. A clear illustration of this claim is the generalization of Central Pattern Generators (CPG), first identified in the cat spinal cord, to generate cyclic motion in robotic locomotion. Despite a global endorsement of this model, physiological and functional experiments in mammals have also indicated the presence of descending signals from the cerebellum, and reflex feedback from the lower limb sensory cells, that closely interact with CPGs. To this day, these interactions are not fully understood. In some studies, it was demonstrated that pure reflex-based locomotion in the absence of oscillatory signals could be achieved in realistic musculoskeletal simulation models or small compliant quadruped robots. At the same time, biological evidence has attested the functional role of the cerebellum for predictive control of balance and stance within mammals. In this paper, we promote both approaches and successfully apply reflex-based dynamic locomotion, coupled with a balance and gravity compensation mechanism, on the state-of-art HyQ robot. We discuss the importance of this stability module to ensure a correct foot lift-off and maintain a reliable gait. The robotic platform is further used to test two different architectural hypotheses inspired by the cerebellum. An analysis of experimental results demonstrates that the most biologically plausible alternative also leads to better results for robust locomotion.
\end{abstract}

\section{Introduction}

The fields of robotics and neuroscience have actively influenced each other. On one side, biology has been the primary source of inspiration for improving the state of the art in robotic control or mechanical design. On the other side, neuroscience has regularly benefited from robotics to conduct in vitro experiments \cite{Floreano2014} and will undoubtedly continue to be used in the future to validate hypotheses on neural architectures, cognitive mechanisms, or social interactions and development. On this basis, robotics research has demonstrated the key function of Central Pattern Generators (CPG) in locomotion \cite{Ijspeert2008}, and corroborated biological studies conducted a century ago \cite{cpg_brown}. However, several factors are still to be clarified regarding, among others, the role of reflex feedback, the function of descending signals from the cortex, or the importance of muscle compliance. Robots like the actively compliant quadruped HyQ \cite{Semini2011}, endowed with a locomotion controller inspired from biological data, are therefore a convenient tool to validate specific hypotheses about the brain.\\

Three major requirements of the central nervous system have been identified by Grillner to achieve robust locomotion in vertebrates \cite{Tern1985}: producing basic rhythmic patterns of flexor and extensor muscles, providing posture and equilibrium control, and enabling adaptation capabilities to react to environmental changes quickly. In the early 1980s, Raibert demonstrated how dynamic locomotion gaits could be achieved in robotics using a few simple, decoupled control laws \cite{Raibert1986}. The core idea was to dissociate the control of the foot trajectory, ideally periodic with a constant amplitude, the control of the speed, requested by the user, and the balance, ensured using inertial corrections. The sharp separation between posture control, dynamically corrected by playing on actuation torques, and foot trajectory, governed by kinematics equations to achieve a constant locomotive cycle, has also been employed in other state-of-art quadruped robots since then \cite{Papadopoulos2002} \cite{hyqconf}.\\

Although the existence of oscillatory CPGs in mammals has been exhibited in various research works, their role has not yet been proven in human locomotion \cite{Minassian2017} and different works have discussed the importance of reflexes to initiate and maintain locomotion. Simulation studies on cat locomotion have indicated that the stability of gait coordination in the presence of external disturbance depends heavily on sensing Ground Reaction Forces (GRFs) in each separate leg, as well as the mechanical coupling between the legs \cite{Ekeberg2005}. Realistic timing between different elements in the reflex-based locomotion model is also required to generate correct behaviors and generalize to different gaits \cite{Murai2010a}. The fundamental role played by the compliant musculoskeletal system has also been emphasized in \cite{Geyer2010}, where an accurate model of the human legs coupled with local muscle reflexes could produce realistic patterns of walking without any form of global control. It has been shown that the model could also handle disturbance and small obstacles by naturally generating recovery foot trajectories \cite{nakamuras_phd}. Pure reflex-based locomotion has also been achieved in robotics where embodied walking, trotting and bounding gaits were implemented on small compliant quadruped robots using only proprioceptive feedback with no timing information \cite{Degrave2015} \cite{Vandesompele2019}.\\

Notwithstanding the ability to model reflex-based gaits using no CPG nor a stability mechanism, the localization of a descending pathway from the brainstem to the spinal cord and its functional role during locomotion has been documented in cats since 1980 \cite{Steeves1980}. The role of the cerebellum is complex and diverse. During locomotion, it has been suggested that this organ implements different functions including the initiation of locomotion patterns \cite{Jordan2008}, the modification of gait and posture on uneven terrain \cite{Drew2004}, or the regulation of interlimb coordination and gait transitions \cite{Danner2016}. Cerebellar lesions on walking cats have demonstrated the active role of the cerebellum by showing abnormal timing of relative limb movements, reduced amplitude in different joints, and decreased stride lengths \cite{Yu1983}. In \cite{Morton2004}, the authors indicate that the medial zone and the flocculonodular lobe in the human cerebellum influence the control of extensor muscles to maintain correct balance and a proper stance and modulate the rhythm of locomotion patterns. Experiments with patients affected by cerebellar ataxia also demonstrated a decreased stability of the trunk's center of mass due to a deteriorated stance in the presence of lateral and backward disturbance during locomotion compared to healthy subjects \cite{Bakker2006}. A discussion of the adaptive role of the cerebellum in locomotion has been provided in \cite{Morton2006}. Experiments on patients with ataxia walking on a split-belt treadmill demonstrated that cerebellar impairment did not decrease reactive feedback-driven adjustments, but significantly damaged predictive feedforward motor adaptations. This evidence supports the hypothesis that the cerebellum helps during locomotion in predicting the limb movements using a stored internal representation with spatial and temporal components.\\

In this paper, we first discuss how robotic reflex-based locomotion in synergy with a stability mechanism can be practically achieved on the actively compliant HyQ robot. To this goal, we train a closed-loop end-to-end neural network to control HyQ in a trotting task. This biologically-inspired architecture takes only the GRFs as inputs and directly outputs the robot's joints position and speed. We show that the dynamic system formed by the neural network and the robot can converge to a stable attractor during treadmill experiments, if supported by a module for gravity compensation and balance control. On a heavy compliant robot, this module turns out to be crucial for providing a good lift-off and a correct swing amplitude, which in turn stabilizes neural control.\\

Secondly, driven by biological evidence concerning the cerebellum, we formulate the hypothesis that this stability mechanism can benefit from a temporal representation of the gait. To that end, we compare two models: one relying on a predictive pattern of desired stance (called PSE) and a second, less biologically plausible, relying on feet position above the ground (called RSE). This experimental design is loosely inspired by the work conducted on humans in \cite{Morton2006}. We demonstrate that only the first model leads to robust locomotion on HyQ, which supports the idea that functional inspiration from the cerebellum can have a positive impact in robotics locomotion.\\

\section{Methods}
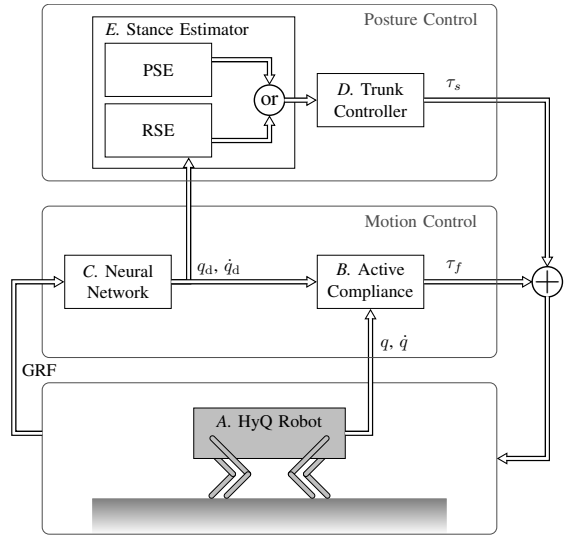
\begin{figure}[thpb]
	\centering
	\resizebox{\columnwidth}{!}{\definecolor{blue538}{RGB}{48, 162, 218}
\definecolor{orange538}{RGB}{252, 79, 48}
\definecolor{yellow538}{RGB}{229, 174, 56}
\definecolor{green538}{RGB}{109, 144, 79}
\definecolor{gray538}{RGB}{139, 139, 139}
\definecolor{grey}{RGB}{50, 50, 50}
\definecolor{violet538}{RGB}{70, 39, 89}

\begin{tikzpicture}[arc/.style={draw,thick,->}]
	
\clip (-1.2, 0) rectangle + (12, 10.5);

\tikzstyle{block} = [draw, rectangle, minimum height=3em, minimum width=6em, align=center, node distance=2.7cm]
\tikzstyle{input} = [coordinate, node distance=1.7cm]
\tikzstyle{vecArrow} = [thick, decoration={markings,mark=at position 1 with {\arrow[semithick]{open triangle 60}}},
double distance=1.4pt, shorten >= 5.4pt,
preaction = {decorate},
postaction = {draw,semithick,line width=1.4pt, white,shorten >= 4.5pt}]
\tikzstyle{lineArrow} = [semithick, decoration={markings,mark=at position 1 with {\arrow{triangle 60}}},
preaction = {decorate},]
\newcommand*\circled[1]{\tikz[baseline=(char.base)]{
		\node[shape=circle,draw,inner sep=2pt] (char) {#1};}}

\begin{scope}

	\draw[gray, fill=gray, path fading=south] (1, 0) rectangle +(7,0.7);
	\draw[rounded corners=0.5mm, black, fill=lightgray, rotate around={-135:(3.78, 1.19)}] (3.735, 1.14) rectangle +(0.7,0.1);
	\draw[rounded corners=0.5mm, black, fill=lightgray, rotate around={-45:(5.22, 1.19)}] (5.175, 1.14) rectangle +(0.7,0.1);
	\draw[rounded corners=0.5mm, black, fill=lightgray, rotate around={-45:(3, 1.9)}] (3, 1.9) rectangle +(1.1,0.1);
	\draw[rounded corners=0.5mm, black, fill=lightgray, rotate around={45:(6, 1.9)}] (6, 1.9) rectangle +(-1.1,0.1);
	\draw[black, fill=lightgray] (3, 1.5) rectangle +(3,1);
	\draw[rounded corners=0.5mm, black, fill=lightgray, rotate around={-45:(3.3, 1.9)}] (3.3, 1.9) rectangle +(1.1,0.1);
	\draw[rounded corners=0.5mm, black, fill=lightgray, rotate around={45:(5.7, 1.9)}] (5.7, 1.9) rectangle +(-1.1,0.1);
	\draw[rounded corners=0.5mm, black, fill=lightgray, rotate around={-135:(4.08, 1.19)}] (4.035, 1.14) rectangle +(0.7,0.1);
	\draw[rounded corners=0.5mm, black, fill=lightgray, rotate around={-45:(4.92, 1.19)}] (4.875, 1.14) rectangle +(0.7,0.1);
	
	\draw [rectangle, color=black!70, rounded corners] (0, 0) rectangle +(9, 3);
	\node[text width=5cm, align=center] at (4.5, 2.2) {\textit{A.} HyQ Robot};
	
	\node [input] (in1) at (0, 5) {};
	\draw node [block, right of=in1, node distance=1.5cm] (nn) {\textit{C.} Neural\\Network};
	\draw node [block, right of=nn, node distance=5cm] (compl) {\textit{B.} Active\\ Compliance};
	\node[text width=2cm, align=center] at (3.5, 5.3) {$q_{\mathrm{d}}$, $\dot{q}_{\mathrm{d}}$};
	\draw[vecArrow] (compl) to (9.71, 5);
	\node[text width=1cm, align=left] at (8.5, 5.3) {$\tau_f$};
	\draw[vecArrow] (6, 2) -| (compl.south);
	\draw [rectangle, color=black!70, rounded corners] (0, 3.5) rectangle +(9, 3);
	\node[text width=1cm, align=left] at (7.2, 3.8) {$q$, $\dot{q}$};
	\node[text width=5cm, align=center, color=black!70] at (7.5, 6.2) {Motion Control};
	
	\node [input] (in2) at (0, 9.2) {};
	\node [input] (in3) at (0, 8.6) {};
	\node [input] (in4) at (0, 8) {};
	\draw node [block, right of=in3, node distance=6.5cm] (tc) {\textit{D.} Trunk\\Controller};
	\draw[black] (1, 7.25) rectangle +(4,3);
	\node[text width=3.5cm, align=left] at (3, 10)  {\textit{E.} Stance Estimator};
	\draw[thick](4.5, 8.6) circle  (0.3cm);
	\node[text width=0.3cm, align=center] (mix) at (4.475, 8.6) {\large or};
	\draw node [block, right of=in4, node distance=2.3cm] (rse) {RSE};
	\draw node [block, right of=in2, node distance=2.3cm] (pse) {PSE};
	\draw [vecArrow] ($(pse.east) + (0, 0.2)$) -| (4.5, 8.9);
	\draw [vecArrow] ($(rse.east) - (0, 0.2)$) -| (4.5, 8.3);
	\draw [vecArrow] (4.8, 8.6) to (tc);
	\draw[vecArrow] (tc) -| (10, 5.29);
	\draw[vecArrow] (nn.east) -| ($(rse.south) + (0.6, 0)$);
	\draw[vecArrow] (nn) to (compl);
	\node[text width=1cm, align=left] at (8.5, 8.9) {$\tau_s$};
	\draw [rectangle, color=black!70, rounded corners] (0, 7) rectangle +(9, 3.5);
	\node[text width=5cm, align=center, color=black!70] at (7.5, 10.2) {Posture Control};
	
	\draw[thick](10, 5) circle  (0.3cm);
	\node[text width=0.3cm, align=center] (mixmain) at (9.89, 5) {\huge $+$};
	\draw[vecArrow] (10, 4.71) |- (9, 1.5);
	\draw[vecArrow] (0, 2) -| ($(nn.west) - (1, 0)$) -- (nn);
	\node[text width=5cm, align=left] at (2.1, 3.25) {GRF};
	
\end{scope}

\end{tikzpicture}}  

	\caption{Diagram of the experimental architecture. The lower level illustrates the HyQ robot on a treadmill. The middle part includes the active compliance module for the different joints and the neural network that controls foot trajectories in closed-loop using GRF feedback. The upper part is a functional model of the cerebellum correcting the stance and the balance of the robot.}
	\label{fig:architecture}
\end{figure}

%
The overall control architecture used in the experiments is presented in Fig. \ref{fig:architecture}. It is divided into three parts: the real robot in its environment, the reflex-based motion controller and the posture controller inspired by the cerebellum. Each component is detailed one by one in the different sections hereafter.\\
%
\subsection{HyQ Robot}
HyQ is a state-of-the-art hydraulically powered quadruped platform of 1.3m length and 90kg weight \cite{Semini2011}. It can walk \cite{hyq_crawl} and trot \cite{hyqconf} robustly on uneven terrains with obstacles of different heights thanks to its visual, torque and inertial sensors and the fast reactivity of its actuators.\\

\begin{figure}[htbp]
\centering
\vspace{3mm}
\includegraphics[width=0.9\columnwidth]{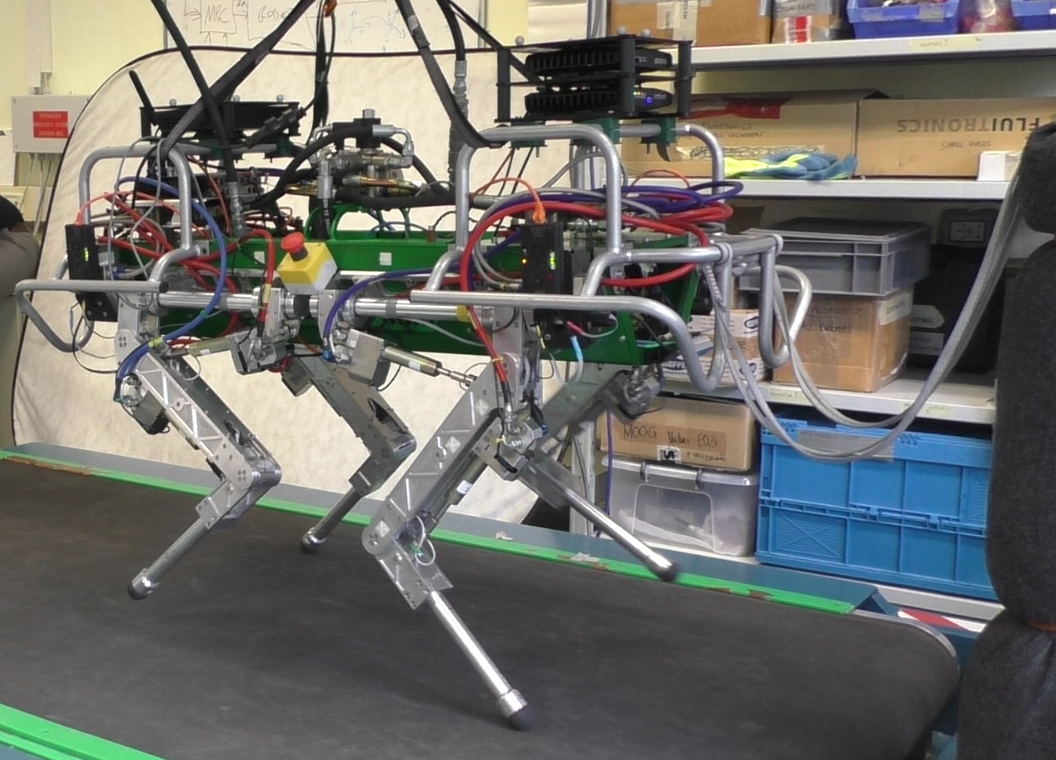}
\caption{HyQ is trotting using reflex-based neural network control.}\label{fig:hyq}
\end{figure}

As presented in Fig. \ref{fig:hyq}, HyQ has four legs with three degrees-of-freedom each, named Hip Abduction-Adduction (HAA), Hip Flexion-Extension (HFE) and Knee Flexion-Extension (KFE). All joints are hydraulically actuated. The advantage of this feature regarding the current work is twofold: first, the joints are capable of delivering or dissipating high torques, which allows fast actuation and makes the robot particularly robust for testing a feed-forward neural network controller, prone to oscillating behaviors that lead to larger GRFs; secondly, the actuation system can virtually produce adjustable levels of damping and stiffness as described in the following section.\\
%
\subsection{Active Compliance Module}
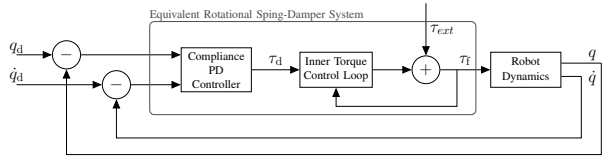
\begin{figure}[htbp]
	\centering
	\resizebox{\columnwidth}{!}{	
	\tikzstyle{block} = [draw, rectangle, minimum height=3em, minimum width=6em]
	\tikzstyle{sum} = [draw, circle]
	\tikzstyle{input} = [coordinate, node distance=1.7cm]
	\tikzstyle{output} = [coordinate, node distance=1.5cm]
	\tikzstyle{pinstyle} = [pin edge={to-,thin,black}]
	
	\newcommand{\suma}{\Large$+$}
	\newcommand{\diffa}{\Large$-$}
	\newcommand{\derv}{\huge$\frac{d}{dt}$}
	
	\begin{tikzpicture}[auto, thick, node distance=2.7cm, >=triangle 45]
	\draw
	node [input] (input1) {} 
	node [input, below of=input1, node distance=0.9cm] (input2) {} 
	node [sum, right of=input1, node distance=1.5cm] (diffa1) {\diffa}
	node [sum] (diffa2) at (3, -0.9) {\diffa}
	node [input, below of= input1, node distance= 0.45cm] (input3) {} 	
	
	node [block, right of=input3, align=center, node distance= 6cm] (pd) {Compliance\\ PD\\Controller}
	node [block, right of=pd, align=center, node distance=3.6cm] (torque) {Inner Torque\\Control Loop}
	node [sum, right of=torque] (suma1) {\suma}
	node [input, above of=suma1, node distance=2cm] (input4) {}
	node [block, right of=suma1, align=center, node distance=3cm] (dyn) {Robot\\Dynamics}
	node [output, right of=dyn, node distance=3cm] (output1) {};
		
	\path[->] (input1) edge node {} (diffa1);
	\path[->] (input2) edge node {} (diffa2);
	\path[->] (diffa1) edge node {} ($(pd.west) +(0, 0.45)$);
	\path[->] (diffa2) edge node {} ($(pd.west) -(0, 0.45)$);
	\path[->] (pd) edge node {\Large$\tau_{\mathrm{d}}$} (torque);
	\path[->] (torque) edge node {} (suma1);
	\path[->] (suma1) edge node {\Large$\tau_{\mathrm{f}}$} (dyn);
	\path[-] ($(dyn.east) +(0, 0.2)$) edge node {} ($(dyn.east) +(1.2, 0.2)$);
	\path[-] ($(dyn.east) -(0, 0.2)$) edge node {} ($(dyn.east) +(0.6, -0.2)$);
	\path[->] (input4) edge node {\Large$\tau_{ext}$} (suma1);
	\draw[->] ($(suma1.east) + (0.5, 0)$) |- ($(suma1.east) -(0, 1.1)$) -| (torque.south);
	\draw[->] ($(dyn.east) +(0.6, -0.2)$) |- (6, -2.5) -| (diffa2.south);
	\draw[->] ($(dyn.east) +(1.2, 0.2)$) |- (6, -3) -| (diffa1.south);
	\node[text width=2cm, align=center] at (1.8, -1.7) {};
	\node[text width=1cm, align=center] at (0, 0.2)  {\Large$q_{\mathrm{d}}$};
	\node[text width=1cm, align=center] at (0, -0.6)  {\Large$\dot{q}_{\mathrm{d}}$};
	\node[text width=1cm, align=center] at (17.3, 0)  {\Large$q$};
	\node[text width=1cm, align=center] at (17.3, -0.7)  {\Large$\dot{q}$};
	
	\draw [rectangle, color=black!70, rounded corners] (4, -1.8) rectangle +(9.8, 2.8);
	\node[text width=8cm, color=black!70, align=left] at (8, 1.2) {Equivalent Rotational Sping-Damper System};

	
	
		
	\end{tikzpicture}}  

	\caption{The torque controller of the leg's actuators allows an accurate reproduction of virtual stiffness and damping properties. The speed of the hydraulic actuation and the fast PD loop is a key factor in its high performance \cite{hyq_hydraulic_compliance_contoller}.}\label{fig:compliance}
\end{figure}

To control the mechanical impedance, we have used the implementation described in \cite{hyq_hydraulic_compliance_contoller}. The PD controller presented in Fig. \ref{fig:compliance} outputs the torque for each actuator and provides them with virtual stiffness and damping. In a first approximation, we assume that the dynamics of the \textit{Inner Torque Control Loop} is negligible compared to the impedance controller sampled at 250Hz. The equation for each joint can be expressed as:
\begin{equation}
\tau_{\mathrm{f}} = \tau_{\mathrm{ext}} + \ K_{\mathrm{p}} \ \big( q_{\mathrm{d}} - q \big) + \ K_{\mathrm{d}} \ \big( \dot{q}_{\mathrm{d}} - \dot{q} \big),
\end{equation}
Where $q_\mathrm{d}$ and $q$ are the desired and actual joint positions; $\tau_f$ is the final torque applied on the joint and $\tau_{ext}$ the eventual disturbance.

The proportional and derivative gains $K_\mathrm{p}$ and $K_\mathrm{d}$ can respectively represent the stiffness $k$ and the damping $c$ of an equivalent rotational spring-damper system of which we would vary the reference angle and rotational speed during actuation. Therefore, $K_\mathrm{p}$ can be measured in N.m/rad and $K_\mathrm{d}$ in N.m.s/rad. The validity of this approach and the assumptions are discussed in \cite{hyq_compliance_stability} and we have based ourselves on these conclusions to select the gains to avoid underdamped responses and ensure stability in our trials.\\\\
%
\subsection{Reflex-based Neural Network}
%
Closed-loop control using a feed-forward neural network has demonstrated some potential to regulate locomotion of complex and compliant systems \cite{Caluwaerts2012, Degrave2015, Urbain2017}. This technique lowers the needs of prior knowledge about the robot’s model and simplifies the control architecture. Flexible and compliant robots, with increased adaptability and energy efficiency, constitute an illustrative category where this feature can be particularly beneficial .\\

\begin{figure}[thpb]
	\centering
	\resizebox{\columnwidth}{!}{	\definecolor{blue538}{RGB}{48, 162, 218}
	\definecolor{orange538}{RGB}{252, 79, 48}
	\definecolor{yellow538}{RGB}{229, 174, 56}
	\definecolor{green538}{RGB}{109, 144, 79}
	\definecolor{gray538}{RGB}{139, 139, 139}
	\definecolor{violet538}{RGB}{70, 39, 89}
	
	\begin{tikzpicture}[arc/.style={draw,thick,->}]
	
	\clip (1, 7) rectangle (15, 11);
	
	\tikzstyle{block} = [draw, rectangle, minimum height=3em, minimum width=6em, align=center, node distance=2.7cm]
	\tikzstyle{input} = [coordinate, node distance=1.7cm]
	\tikzstyle{vecArrow} = [thick, decoration={markings,mark=at position 1 with {\arrow[semithick]{open triangle 60}}},
	double distance=1.4pt, shorten >= 5.4pt,
	preaction = {decorate},
	postaction = {draw,semithick,line width=1.4pt, white,shorten >= 4.5pt}]
	\tikzstyle{vecArrowOrange} = [thick,orange538, decoration={markings,mark=at position 1 with {\arrow[semithick, orange538]{open triangle 60}}},
	double distance=1.4pt, shorten >= 5.4pt,
	preaction = {decorate},
	postaction = {draw,semithick,line width=1.4pt, white,shorten >= 4.5pt}]
	\tikzstyle{vecArrowGreen} = [thick,green538, decoration={markings,mark=at position 1 with {\arrow[semithick, green538]{open triangle 60}}},
	double distance=1.4pt, shorten >= 5.4pt,
	preaction = {decorate},
	postaction = {draw,semithick,line width=1.4pt, white,shorten >= 4.5pt}]
	\tikzstyle{innerWhite} = [semithick, white,line width=1.4pt, shorten >= 4.5pt]

	\newcommand\timebuffer{
		\begin{tikzpicture}
		\draw [rectangle, align=center] (0, 0) rectangle +(3, 0.5);
		\draw (0.5,0) -- (0.5,0.5);
		\draw (1,0) -- (1,0.5);
		\draw (2.5,0) -- (2.5,0.5);
		\node[text width=0.5cm, align=center] at (0.25, 0.25) {1};
		\node[text width=0.5cm, align=center] at (0.75, 0.25) {2};
		\node[text width=1cm, align=center] at (1.75, 0.25) {...};
		\node[text width=0.5cm, align=center] at (2.75, 0.25) {\small $M$};
		\node[text width=2cm, align=center] at (1.5, -0.3) {Time Buffer};
		\node[text width=2cm, align=center] at (1.5, 1.1) {};
		\end{tikzpicture}
	}
	
	\newcommand\elm{
		\begin{tikzpicture}
		\draw [rectangle, align=center] (0, 0) rectangle +(0.5, 2.5);
		\draw (0.25, 0.25) circle  (0.2cm);
		\node[text width=0.3cm, align=center] at (0.25, 0.22) {$\sim$};
		\draw (0.25, 2.25) circle  (0.2cm);
		\node[text width=0.3cm, align=center] at (0.25, 2.22) {$\sim$};
		\draw (0.25, 1.75) circle  (0.2cm);
		\node[text width=0.3cm, align=center] at (0.25, 1.72) {$\sim$};
		\node[text width=0.5cm, align=center] at (0.75, 2.25) {1};
		\node[text width=0.5cm, align=center] at (0.75, 1.75) {2};
		\node[text width=0.5cm, align=center] at (0.25, 1) {...};
		\node[text width=0.5cm, align=center] at (0.75, 0.25) {\small $N$};
		\node[text width=5cm, align=center] at (0.25, -0.8) {ELM};
		\node[text width=2cm, align=center] at (0, -2.7) {};
		\end{tikzpicture}
	}
	
	\newcommand\nn{
		\begin{tikzpicture}
		\draw (0, 0) circle  (0.15cm);
		\node[text width=0.3cm, align=center] at (-0.012, 0.0) {\large $\times$};
		\draw[->] (0.15, 0) -- (0.5, 0);
		\draw (0, 0.35) circle  (0.15cm);
		\node[text width=0.3cm, align=center] at (-0.012, 0.35) {\large $\times$};
		\draw[->] (0.15, 0.35) -- (0.5, 00.35);
		\draw (0, 0.7) circle  (0.15cm);
		\node[text width=0.3cm, align=center] at (-0.012, 0.7) {\large $\times$};
		\draw[->] (0.15, 0.7) -- (0.5, 0.7);
		\draw (0, 1.05) circle  (0.15cm);
		\node[text width=0.3cm, align=center] at (-0.012, 1.05) {\large $\times$};
		\draw[->] (0.15, 1.05) -- (0.5, 1.05);
		\draw (0, 1.4) circle  (0.15cm);
		\node[text width=0.3cm, align=center] at (-0.012, 1.4) {\large $\times$};
		\draw[->] (0.15, 1.4) -- (0.5, 1.4);
		\draw (0, 1.75) circle  (0.15cm);
		\node[text width=0.3cm, align=center] at (-0.012, 1.75) {\large $\times$};
		\draw[->] (0.15, 1.75) -- (0.5, 1.75);
		\draw (0, 2.1) circle  (0.15cm);
		\node[text width=0.3cm, align=center] at (-0.012, 2.1) {\large $\times$};
		\draw[->] (0.15, 2.1) -- (0.5, 2.1);
		\draw (0, 2.45) circle  (0.15cm);
		\node[text width=0.3cm, align=center] at (-0.012, 2.45) {\large $\times$};
		\draw[->] (0.15, 2.45) -- (0.5, 2.45);
		\node[text width=4cm, align=center] at (0, -0.8) {Linear\\Readout Layer};
		\node[text width=1cm, align=center] at (0, -2.85) {};
		\end{tikzpicture}
	}
	
	\draw node [input] (in5) at (0, 8) {};
	\draw node [block, right of=in5, node distance=3cm] (scaler) {MinMax Scaler\\$[-1.5; 1.5]$};
	\draw node [right of=scaler, node distance=4cm] (buff) {\timebuffer};
	\draw node [right of=buff, node distance=4cm] (elm) {\elm};
	\draw node [right of=elm, node distance=2.5cm] (nn) {\nn};
	
	\draw[vecArrow] (in5) -- (scaler);
	\draw[vecArrow] (scaler) -- ($(buff.west)+(6.5pt,0)$);

	\foreach [count=\i] \buffx in {-1.25, -0.75,  1.25}{
		\foreach [count=\j] \elmy in {2.4, 1.9, 1.4, 0.4}{	
			\draw[color=gray!60] ($(buff.north)+(\buffx, -0.825)$) -- ($(elm.west)+(2.5, \elmy)$);
		}
	}
	
	\foreach [count=\i] \elmy in {2.4, 1.9, 1.4, 0.4}{
		\foreach [count=\j] \nny in {0.13, 0.48, 0.83, 1.18, 1.53, 1.88, 2.23, 2.58}{	
			\draw[color=blue538!60] ($(elm.east)+(-2.49, \elmy)$) -- ($(nn.west)+(2.08, \nny)$);
		}
	}
	

	\end{tikzpicture}}  

	\caption{The architecture of the neural network. A time buffer and a hidden layer of hyperbolic tangent neurons allow us to easily tune the memory and non-linearity parameters. The time buffer is fully connected to the hidden layer and only the connections between the hidden layer and the linear readout are trained (blue).}\label{fig:nn}
\end{figure}
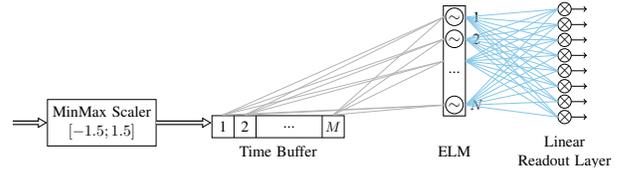

The architecture of the reflex-based neural network used in the trials is presented in Fig. \ref{fig:nn}. It receives four scalar GRFs from the feet as inputs and it outputs eight positions and eight speeds for the HFE and KFE joints of all legs. The inputs are first normalized, then sent to a time buffer, which acts as a \textit{first-in, first-out} queue. It is fully connected to a hidden layer of hyperbolic tangent neurons, followed by another fully connected readout layer of linear neurons. The layers architecture is inspired by the Extreme Learning Machines (ELM) \cite{ELM} and has been successfully used in \cite{jonas_elm} before. This feed-forward architecture displays only two parameters, $M$ and $N$, to tune the memory and the nonlinear hidden projections of the controller's model, respectively. In this paper, we fix both $M$ and $N$ to 80. This choice guarantees accurate and stable predictions in closed-loop. A discussion on the subject is out of the scope of this article and is carried in \cite{hyq_under_review}. This study also investigates more scrupulously the influence of the body parameters on the complexity of the neural network and the locomotion performance.\\

The chronology of each experimental trial is divided into three phases. In the first phase of 120 seconds, the weights of the connections between the hidden and the linear readout layer are trained using the FORCE learning method \cite{Sussillo} to learn to reproduce target cyclic feet trajectories produced by the HyQ Reactive Controller Framework (RCF) \cite{hyqconf}. In this phase, the robot only uses the target signal to trot. In the second phase of 30 seconds, we switch from the target to the predicted signals and the algorithm progressively alternates between training and prediction modes until it fully works in prediction at the end of the phase. The final phase also lasts 30 seconds and is dedicated to testing. In this phase, the robot is completely controlled by the architecture presented in Fig. \ref{fig:architecture} and the data is recorded for further analysis.

\subsection{Trunk Controller}
The trunk controller module is part of the RCF controller \cite{hyqconf}. It performs robot stabilization by providing the robot's body (or trunk) with force to correct its pose and compensate gravity. Using the joint and feet velocities, the trunk controller maps the desired body forces into joint torques without moving the feet position in the world’s horizontal plane. To compute the torques, the module also requires information on the stance status, i.e., which leg is in contact with the ground (stance phase) or in the air (swing phase). In the trials presented in the following sections, the different gains in this module are directly extracted from the previous work conducted on HyQ with the RCF controller.

\subsection{Stance Estimator}
\begin{figure}[thpb]
	\begin{flushleft}
	\resizebox{\columnwidth}{!}{\definecolor{blue538}{RGB}{48, 162, 218}
\definecolor{orange538}{RGB}{252, 79, 48}
\definecolor{yellow538}{RGB}{229, 174, 56}
\definecolor{green538}{RGB}{109, 144, 79}
\definecolor{gray538}{RGB}{139, 139, 139}
\definecolor{grey}{RGB}{50, 50, 50}
\definecolor{violet538}{RGB}{70, 39, 89}

\begin{tikzpicture}
	
\begin{scope}

\draw[->] (-0.005,0) -- (6,0) node[below] {};
\draw[-] (0,-0.005) -- (0,2.4) node[above] {};
\def\ynames{{"Right-hind", "Left-hind", "Left-front", "Right-front"}}
\def\ycoord{{0.2, 0.8, 1.4, 2.0}}
\foreach \i in {0, 1,..., 3} {
\node[text width=2cm, align=right] at (-1.3, \ycoord[\i]) {\pgfmathparse{\ynames[\i]}\pgfmathresult};
}
\node[text width=4cm, align=center] at (6, -0.3) {t};

\draw [rectangle, fill=black!40] (0, 0.) rectangle +(1.2, 0.55);
\draw [rectangle, fill=black!60] (0, 1.2) rectangle +(1.2, 0.55);
\draw [rectangle, fill=black!80] (0.8, 0.6) rectangle +(2.4, 0.55);
\draw [rectangle, fill=black!20] (0.8, 1.8) rectangle +(2.4, 0.55);
\draw [rectangle, fill=black!40] (2.8, 0) rectangle +(2.4, 0.55);
\draw [rectangle, fill=black!60] (2.8, 1.2) rectangle +(2.4, 0.55);
\draw [rectangle, fill=black!80] (4.8, 0.6) rectangle +(1.2, 0.55);
\draw [rectangle, fill=black!20] (4.8, 1.8) rectangle +(1.2, 0.55);

\end{scope}

\end{tikzpicture}}  

	\end{flushleft}
	\caption{Trotting pattern for the four legs of HyQ. Each bar represents a leg stance phase and each blank a swing phase}\label{fig:pattern}
\end{figure}
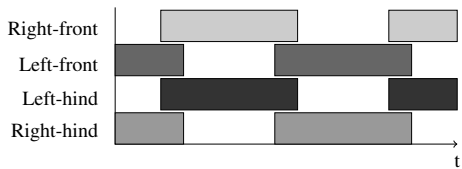

The neural network is trained to predict the foot trajectories for the trotting pattern presented in Fig. \ref{fig:pattern}. In this figure, each bar corresponds to stance phase and each blank represents swing phase. This information is essential to correct the pose of the robot with the trunk controller since this module uses the legs in contact with the ground to produce stabilizing reaction forces. In this paper, we suggest two different stance estimator models that we will refer to as \textit{Predictive Stance Estimator} (PSE) and \textit{Reactive Stance Estimator} (RSE).\\

In the PSE model, the stance is exclusively defined using the desired gait patterns of Fig. \ref{fig:pattern}. We use the RCF controller \cite{hyqconf} to provide this sequence. In our trial configuration, this module generates 3D trajectories depending only on the time variable. These desired trajectories are subsequently thresholded to determine if each foot is in swing or stance mode. As a result, we can write the following simplified equation:
\begin{equation}
\mathrm{PSE} = \text{\large $f$}(t).  \label{eq:des_stance}
\end{equation}
As inspired by biological evidence, this cerebellum function is entirely predictive and integrates a spatiotemporal representation of the desired gait with no reactive feedback induced by the external environment and disturbances.\\

In RSE, however, we rely on feedback from the lower neural network. As explained in the introduction section, this model is not biologically plausible and builds on a reactive pathway where the desired posture is communicated from the lower spinal neural network to the cerebellum. However, from an engineering point of view, the resulting architecture has the advantage to remove all dependencies to the target patterns after training. In practice, stance/swing information is computed using the vertical component of the desired foot positions predicted by the neural network:
\begin{equation}
\mathrm{RSE} = \text{\large $f$}(z_{\mathrm{FR}},\ z_{\mathrm{FL}},\ z_{\mathrm{HL}},\ z_{\mathrm{HR}}),
\end{equation}
where $z$ is the height of the foot and indexes correspond to the four different legs. The function $f$ evaluates the vertical distance between the foot and the ground. To discriminate between stance and swing modes and eventually cope with potential oscillations, it also includes a threshold coupled with a moving average, whose parameters are tuned heuristically until they correspond to the qualitative robot behavior. This dependency can also be simplified to the joint positions and velocities predicted by the neural network after applying direct kinematics:
\begin{equation}
\mathrm{RSE} = \text{\large $f$}(q_{\mathrm{d}}, \dot{q}_{\mathrm{d}}). \ \label{eq:grf_stance}
\end{equation}\\

\section{Results}

We conducted twelve experimental trials divided into two categories of six trials each. The stance status was computed using PSE in the first category, and RSE in the second. In both cases, a disturbance was applied around $t = 160 \mathrm{s}$ in the training sequence defined in the methods section, using a small delay between the neural network and the motors. Some experiments were realized on a treadmill with different robot forward speeds and some were conducted on the ground with the robot trotting in place. These forward speeds were, however, constant for the whole duration of each trial. Selecting different forward speeds modified the joint trajectories slightly but it did not seem to affect the results in any way, and we do not consider it further in this analysis. A video is provided in Supplementary Material to demonstrate the robot gaits on the treadmill qualitatively.\\

\subsection{Stability of the Limit Cycle}
\begin{figure}[htbp]
	\begin{center}
		\includegraphics[width=\linewidth]{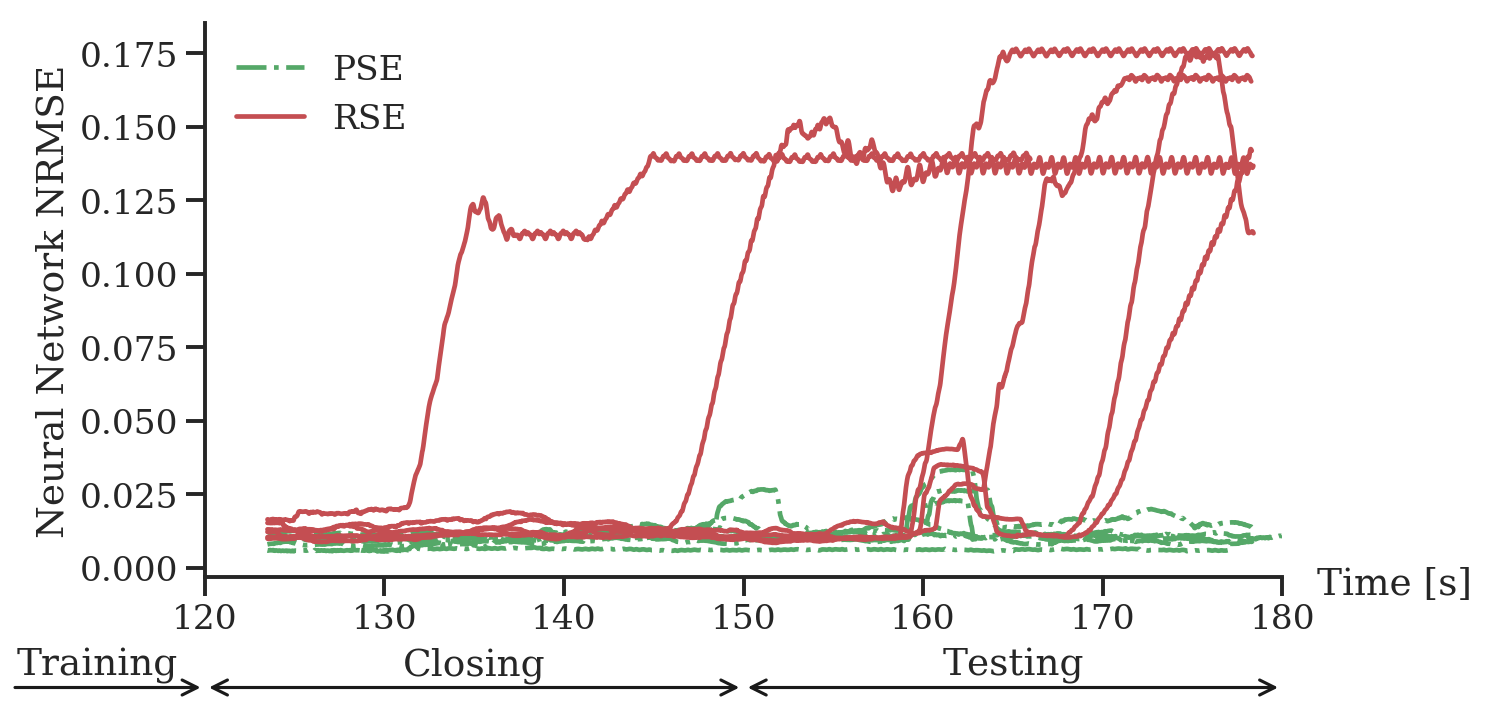}
	\end{center}
	\caption{NRMSE of neural predictions for all experiments, smoothened with a moving average. The trials with a stability module using the RSE model (red) are diverging during the closing or testing phase but not with the PSE model (green).}\label{fig:stability}
\end{figure}

Among the twelve trials, all those using the PSE model succeeded in reproducing the target gait correctly. In contrast, five out of six trials relying on RSE failed in finding a robust limit cycle attractor, which means that the robot eventually ended up falling before the end of the testing phase. These results are displayed in Fig. \ref{fig:stability}. In this graph, we show the Normalized Root Mean Squared Error (NRMSE) between the target and the neural network prediction. A moving average with a window size equal to the gait period is applied on the curve to highlight the global trend. A low NRMSE stands for a good prediction of the joints position and speed, which can reproduce the results of the target RCF controller. The NRMSE is clearly overshooting at different moments in time during the closing and testing phase for the controller models using RSE. Except for the last one, discussed in the next section, they indicate a divergence from the locomotion limit cycle, causing the robot to fall or to act chaotically before we needed to press the emergency stop button. We can also notice that this divergence does not happen only at the time where the disturbance is applied ($t = 160 \mathrm{s}$) but also before or after it, in reaction to the unpredictability of the environment.\\

\begin{figure}[htbp]
\begin{center}
	\subfloat[limitcycle_c][Typical limit cycle with PSE]{\includegraphics[width=0.48\linewidth]{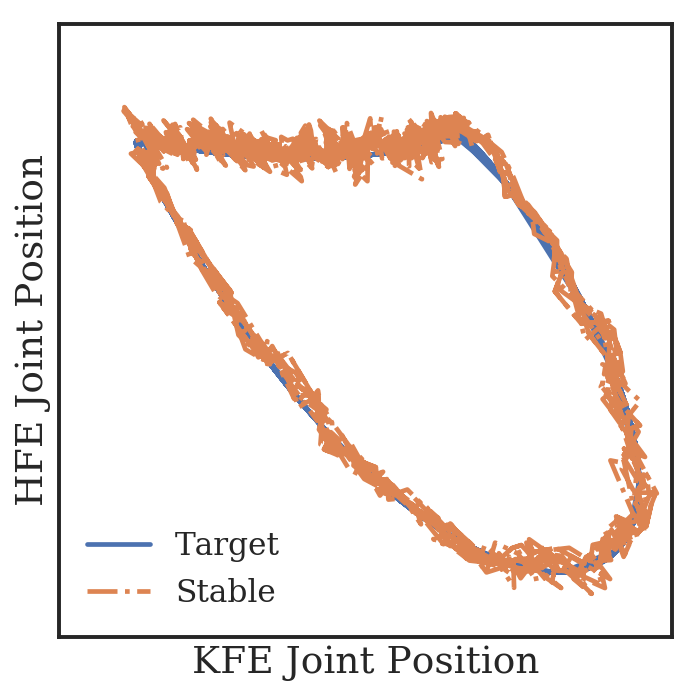}\label{fig:good_attractor}}
	\hfill
	\subfloat[limitcycle_b][Typical limit cycle with RSE]{\includegraphics[width=0.48\linewidth]{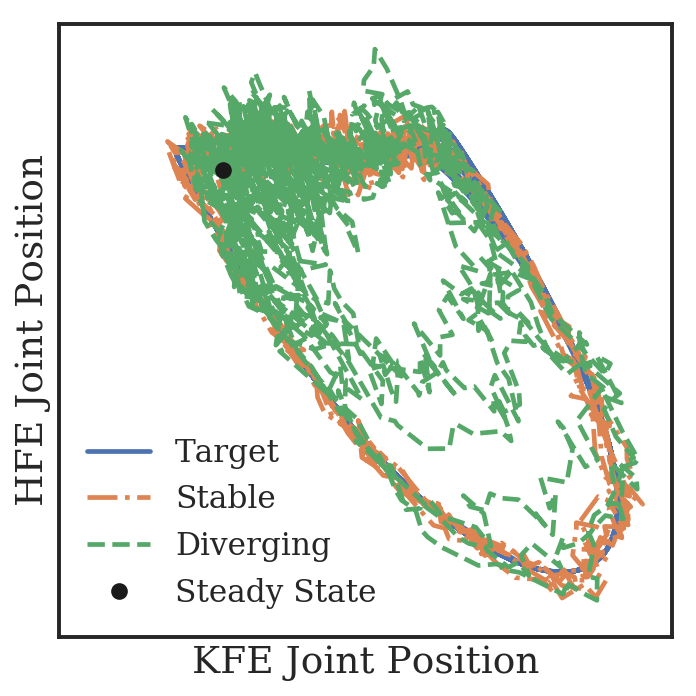}\label{fig:divergent_attractor}}
\end{center}
\caption{Limit cycles for both models. On the left-hand side, the PSE model (orange) is stable and follows correctly the target cycle (blue). On the right-hand side, the RSE model starts from an attractor (orange), close to the target signal (blue), then diverges (green) until it reaches a steady-state point (black).}\label{fig:attractors}
\end{figure}

To further demonstrate these observations, we illustrate a typical limit cycle for trials in both categories in Fig. \ref{fig:attractors}. The instability in green represented in Fig. \ref{fig:divergent_attractor} shows how the limit cycle amplitude decreases until convergence to a steady-state. Such a point is reached when the robot stabilizes, the position of its feet becomes constant (by falling or standing still on the ground) and the cerebellum does not trigger a forced alternation in the stance pattern.\\

%
\subsection{Synchronicity}
\begin{figure}[htbp]
	\begin{center}
		\vspace{1mm}
		\subfloat[synchronicity_b][The control signal of the left-front HFE joint with PSE model]{\includegraphics[width=\linewidth]{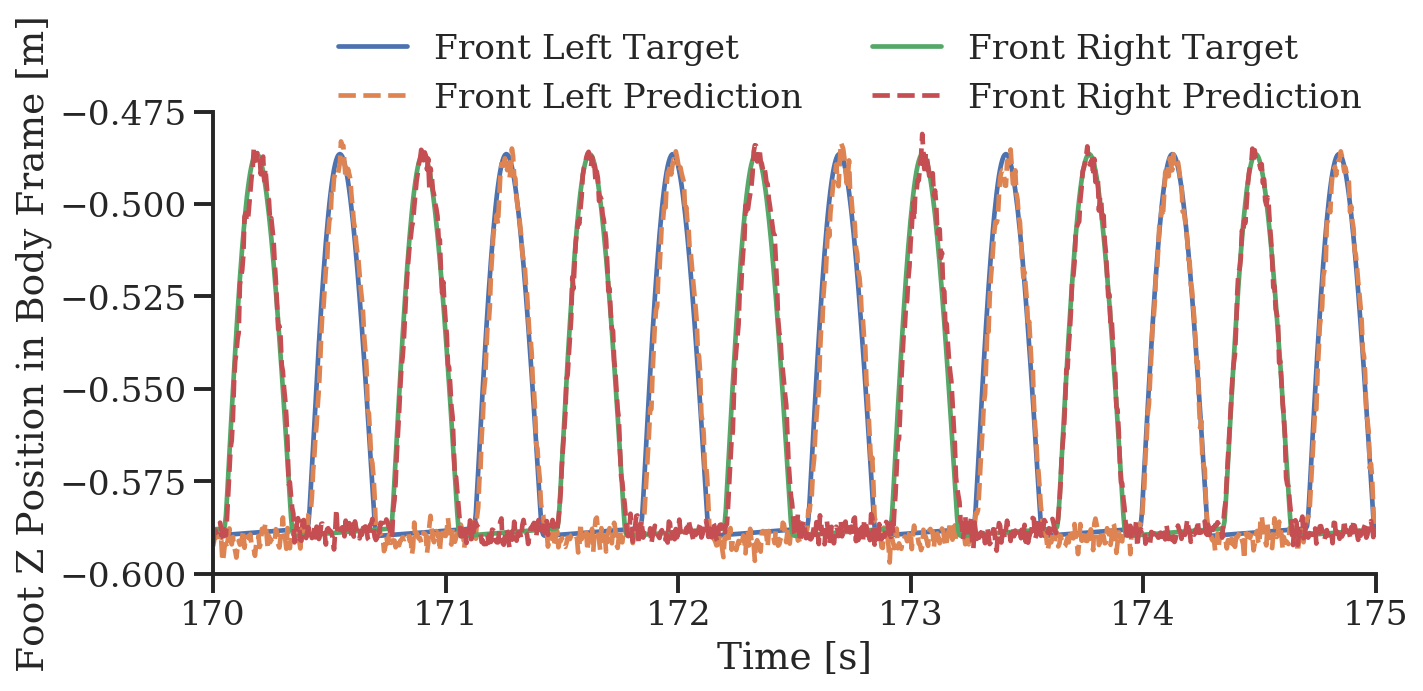}\label{fig:pred_sync}}
		\hfill
		\subfloat[syncrhonicity_a][The control signal of the left-front HFE joint with RSE model]{\includegraphics[width=\linewidth]{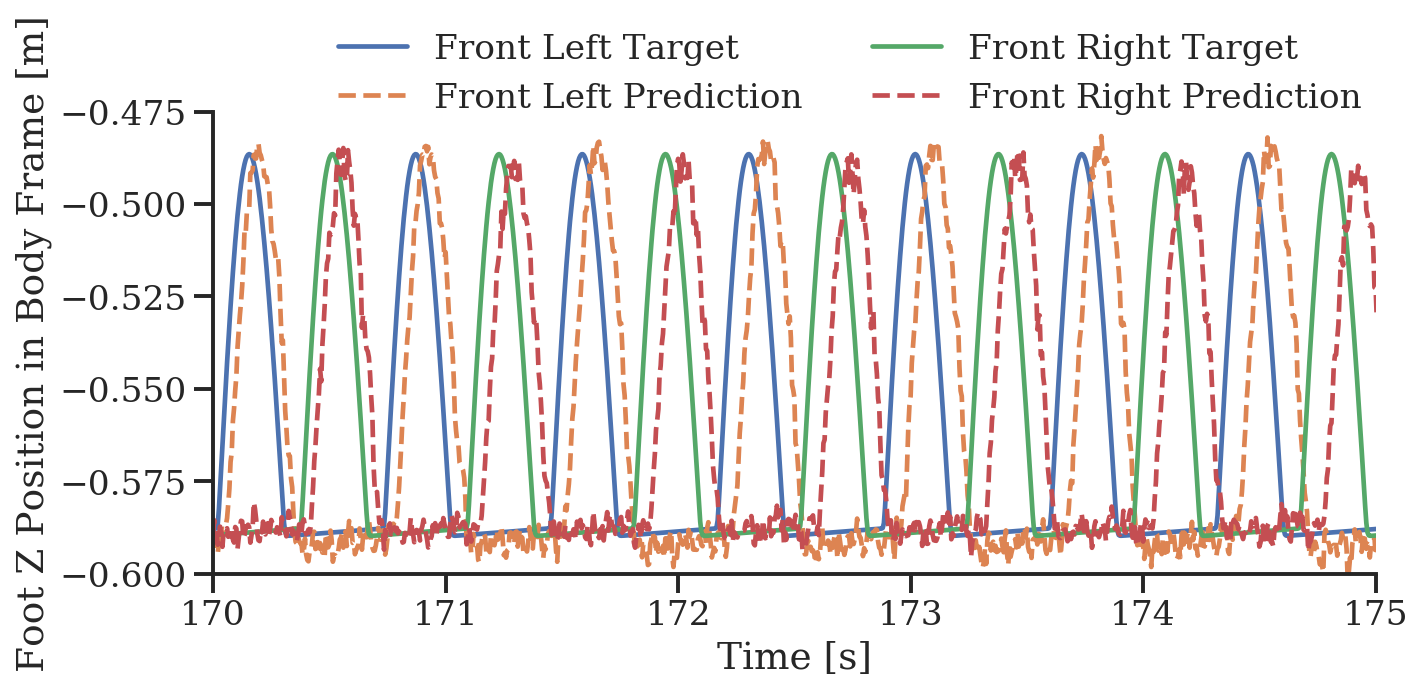}\label{fig:pred_async}}
	\end{center}
	\caption{Quality of frequency-locking with PSE (top) and RSE (bottom). In both graphs, the neural prediction for the height of the left-front (orange) and right-front (red) feet is displayed in comparison with the target signal (blue and green). In the lower graph, the phase is not locked and the gait frequency decreases with time. This does not happen in the upper graph.}\label{fig:synchronicity}
\end{figure}

We mention that the red curve on the far right displayed in Fig. \ref{fig:stability} does not suggest a failure. A visual inspection of the robot behavior during the trial shows that it does not correspond to a chaotic or freezing behavior but a progressive loss of the target signal's phase. In Fig. \ref{fig:pred_async}, we plotted the vertical position of the front feet (for both target and neural prediction) at the end of this trial. The target trajectory has a step height of 10 cm. In comparison, the same signals are represented for a successful trial of the PSE category in Fig. \ref{fig:pred_sync}. The absence of timing information from the cerebellum action leads to a frequency decrease, characterized by a slower robot gait. In contrast, the implicit temporal information embedded in the stability mechanisms in Fig. \ref{fig:pred_sync} ensures a phase-locking on the stance phase, resulting in a neural network prediction synchronized with the target.\\

\section{Discussion}

In this paper, we suggested a locomotion control architecture, inspired by the brain, that includes a cerebellar model for stability and a closed-loop neural network to generate leg movements based on GRF feedback. To interact with the ground and correct the balance, the stability module requires information about which legs are in stance or swing phase. Inspired by evidence from biology, we compared two models to acquire this information: one relying on a predictive pattern of desired stance (PSE) and a second relying on the desired foot positions position along their vertical component (RSE). We formulated the hypothesis that the first model should lead to more stable results as biological studies with patients affected by cerebellar ataxia concluded that a predictive notion of the leg posture is required to improve locomotion stability \cite{Morton2006}. Experiments with both models were conducted on the active compliant quadruped robot HyQ.\\

From the results, we conclude that reflex-based actuation using GRFs can be achieved on a large robot, with twelve degrees of freedom. However, our investigations demonstrate the need for a mechanism of stability and gravity compensation to handle this task. Successful applications in robot locomotion using neural networks generally rely on position control either of stiff robots \cite{biped_walk_dl}, \cite{quadruped_walk_dl}, either of small and light compliant robot \cite{Vandesompele2019}. On a heavy torque-controlled compliant robot, in contrast, the lift-off cannot be easily guaranteed in the absence of a gravity compensation mechanism. In other words, the inherent flexibility of the leg joints and the balancing of the robot body can cause the foot to stay on the ground during the desired swing phase. This effect can have a dramatic impact. First, because the vertical amplitude of the lift-off is crucial to avoid tripping on rough terrain or in the presence of obstacles. Secondly, a stable limit cycle entirely depends on the sequence of contact with the ground when using the biologically-inspired neural network with GRF inputs that we suggested. Therefore, we believe that the architecture presented in this paper can bring a contribution to tackle neural control of locomotion.\\

In a second phase, an analysis of the limit cycle stability pointed out that robust trotting gait can be conducted when using a predictive stance (PSE) in the stability module. From a dynamic point of view, this indicates that reflex-based locomotion requires a timing input to stabilize its limit cycle. In other words, the model seems quite sensible to phase jitter and needs to be corrected with a clock signal. In the single RSE trial where the robot did not fall, the gait frequency could not be held reliably and the robot started to slow down with respect to the required frequency. The same explanation can be used to clarify this effect: in closed-loop and without the desired timing pattern coming from the cerebellum model, the system dynamics are determined by the neural connections and the interaction of the robot with its environment. Undesired external delays will accumulate and frequency locking cannot be guaranteed. In an extreme case, this can lead to robots that slow down until they completely stop and their limit cycle converges to a steady-state. It can also have a disastrous effect if the phase of the different motor commands does not evolve in synchrony, leading to chaotic behavior and falling.\\

The experimental results obtained with our robotic models display a fair correlation with biological observations. First, they fit physiological and functional insights about the cerebellum to work as a predictive circuit, relying on vestibular senses and signals from the cortex but not from the lower limb sensory feedback \cite{Morton2006}. Secondly, they emphasize the role of a clock signal \cite{ivry1997} to achieve robust locomotion. Third, they have a negative effect resulting in frequency loss and larger oscillations of the center of mass, which relates to observations conducted on people with cerebellar ataxia \cite{Bakker2006}.\\

In conclusion, this paper reminds the importance of stability control in neural network control on compliant legged robot locomotion. In particular, it shows how a cerebellum-inspired PSE mechanism ensures a good lift-off and increases robustness to external disturbance and phase shifting. It is important to note that this work on quadruped locomotion directly inspires from a functional model described in human experiments, where the presence of CPG has not been confirmed \cite{Minassian2017}. A better comprehension of how locomotion control evolved from quadrupeds to bipeds should integrate spinal CPG in the model. Such an architecture could also help to clarify the mixed role of CPG and cerebellum to regulate the temporal sequencing in locomotion and should be investigated in future works.

\section*{Acknowledgements}

This research has received funding from the European Union's Horizon 2020 Framework Programme for Research and Innovation under the Specific Grant Agreement No. 785907 (Human Brain Project SGA2). Experiments were conducted on the HyQ Platform hosted and funded by the Fondazione Istituto Italiano di Tecnologia. The authors would also like to thank Shamel Fahmi for his help during the experiments and to process data.

\bibliographystyle{IEEEtran}
\bibliography{biblio}

\end{document}